\documentclass[letterpaper, 10 pt, conference]{ieeeconf}  

\IEEEoverridecommandlockouts                              

\let\labelindent\relax

\usepackage[bookmarks=true]{hyperref}
\usepackage{amsmath}
\usepackage{amssymb}
\usepackage{color}
\usepackage{graphicx}
\usepackage{multirow}
\graphicspath{{images/}}

\usepackage{caption}
\usepackage{subfigure}

\usepackage{setspace} 

\usepackage{amsmath}
\usepackage{amsfonts}
\usepackage{amssymb}
\usepackage{amsthm}
\usepackage{bm}
\usepackage{bbm}
\usepackage{mathtools}
\usepackage{enumitem}
\usepackage{thmtools,thm-restate}
\usepackage{algorithm}
\usepackage{algorithmic}
\usepackage{color}
\usepackage{graphicx}
\usepackage{comment}

\def\*{\star}

\newcommand{\tr}[1]{ \mathrm{tr}\left( #1\right)}

\usepackage[dvipsnames]{xcolor}

\renewcommand{\tr}{{T}}

\newcommand{\Lag}{\mathcal{L}}

\newcommand{\q}{\mathbf{q}}
\newcommand{\qd}{{\dot{\q}}}
\newcommand{\qdd}{{\ddot{\q}}}

\newcommand{\x}{\mathbf{x}}
\newcommand{\xd}{{\dot{\x}}}
\newcommand{\xdd}{{\ddot{\x}}}

\newcommand{\f}{\mathbf{f}}
\newcommand{\h}{\mathbf{h}}

\newcommand{\J}{\mathbf{J}}

\newcommand{\B}{\mathbf{B}}

\newcommand{\I}{\mathbf{I}}

\newcommand{\M}{\mathbf{M}}

\newcommand{\wt}[1]{{\widetilde{#1}}}

\usepackage{amsmath}  
\usepackage{amssymb}
\usepackage{accents}  
\usepackage{amsthm}

\theoremstyle{plain}

\theoremstyle{definition}

\theoremstyle{remark}

\makeatletter
\let\save@mathaccent\mathaccent
\newcommand*\if@single[3]{%
  \setbox0\hbox{${\mathaccent"0362{#1}}^H$}%
  \setbox2\hbox{${\mathaccent"0362{\kern0pt#1}}^H$}%
  \ifdim\ht0=\ht2 #3\else #2\fi
  }
\newcommand*\rel@kern[1]{\kern#1\dimexpr\macc@kerna}
\newcommand*\widebar[1]{\@ifnextchar^{{\wide@bar{#1}{0}}}{\wide@bar{#1}{1}}}
\newcommand*\wide@bar[2]{\if@single{#1}{\wide@bar@{#1}{#2}{1}}{\wide@bar@{#1}{#2}{2}}}
\newcommand*\wide@bar@[3]{%
  \begingroup
  \def\mathaccent##1##2{%
    \let\mathaccent\save@mathaccent
    \if#32 \let\macc@nucleus\first@char \fi
    \setbox\z@\hbox{$\macc@style{\macc@nucleus}_{}$}%
    \setbox\tw@\hbox{$\macc@style{\macc@nucleus}{}_{}$}%
    \dimen@\wd\tw@
    \advance\dimen@-\wd\z@
    \divide\dimen@ 3
    \@tempdima\wd\tw@
    \advance\@tempdima-\scriptspace
    \divide\@tempdima 10
    \advance\dimen@-\@tempdima
    \ifdim\dimen@>\z@ \dimen@0pt\fi
    \rel@kern{0.6}\kern-\dimen@
    \if#31
      \overline{\rel@kern{-0.6}\kern\dimen@\macc@nucleus\rel@kern{0.4}\kern\dimen@}%
      \advance\dimen@0.4\dimexpr\macc@kerna
      \let\final@kern#2%
      \ifdim\dimen@<\z@ \let\final@kern1\fi
      \if\final@kern1 \kern-\dimen@\fi
    \else
      \overline{\rel@kern{-0.6}\kern\dimen@#1}%
    \fi
  }%
  \macc@depth\@ne
  \let\math@bgroup\@empty \let\math@egroup\macc@set@skewchar
  \mathsurround\z@ \frozen@everymath{\mathgroup\macc@group\relax}%
  \macc@set@skewchar\relax
  \let\mathaccentV\macc@nested@a
  \if#31
    \macc@nested@a\relax111{#1}%
  \else
    \def\gobble@till@marker##1\endmarker{}%
    \futurelet\first@char\gobble@till@marker#1\endmarker
    \ifcat\noexpand\first@char A\else
      \def\first@char{}%
    \fi
    \macc@nested@a\relax111{\first@char}%
  \fi
  \endgroup
}
\makeatother

\title{Geometric Fabrics: a Safe Guiding Medium for Policy Learning
}

\author{Karl Van Wyk$^{1}$, Ankur Handa$^{1}$, Viktor Makoviychuk$^{1}$, Yijie Guo$^{1}$, Arthur Allshire$^{1,2}$ and Nathan D. Ratliff$^{1}$
\thanks{$^{1}$Karl Van Wyk, Ankur Handa, Viktor Makoviychuk, Yijie Guo, and Arthur Allshire, and Nathan Ratliff are with Nvidia, USA
        {\tt\footnotesize \{kvanwyk,ahanda\}@nvidia.com}}
\thanks{$^{2}$Arthur Allshire is with University of Toronto, CA}
}

\begin{document}

\makeatletter
\let\@oldmaketitle\@maketitle
\renewcommand{\@maketitle}{\@oldmaketitle
  \includegraphics[width=0.999\linewidth]
    {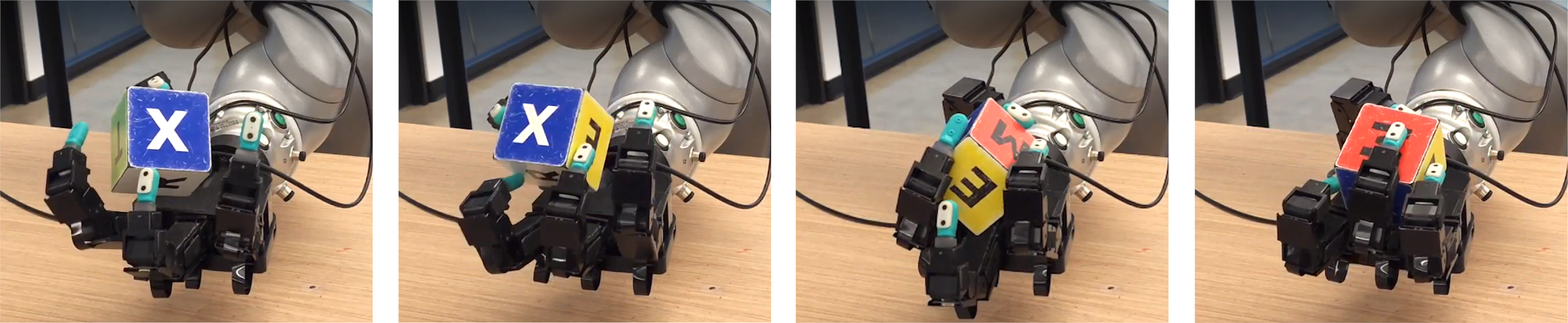} \\
  \refstepcounter{figure}\footnotesize{Fig.~\thefigure. Reinforcement learning over a geometric fabric layer yields safe, high-performance manipulation behavior for a highly-actuated hand. The learned behavior switches between two-, three-, and four-fingered grasps during prehensile manipulation. Videos at \url{https://dextreme.org/fgp.html}.}
  \label{fig:banner_figure} \medskip \vspace{-10pt}}
\makeatother

\maketitle

\begin{abstract}
Robotics policies are always subjected to complex,
second order dynamics that entangle their actions with resulting states. In reinforcement learning (RL) contexts, policies have the burden of deciphering these complicated interactions over massive amounts of experience and complex reward functions to learn how to accomplish tasks. Moreover, policies typically issue actions directly to controllers like Operational Space Control (OSC) or joint PD control, which induces straightline motion towards these action targets in task or joint space. However, straightline motion in these spaces for the most part do not capture the rich, nonlinear behavior our robots need to exhibit, shifting the burden of discovering these behaviors more completely to the agent. Unlike these simpler controllers, geometric fabrics capture a much richer and desirable set of behaviors via artificial, second order dynamics grounded in nonlinear geometry. These artificial dynamics shift the uncontrolled dynamics of a robot via an appropriate control law to form \textit{behavioral dynamics}. Behavioral dynamics unlock a new action space and safe, guiding behavior over which RL policies are trained. Behavioral dynamics enable bang-bang-like RL policy actions that are still safe for real robots, simplify reward engineering, and help sequence real-world, high-performance policies. We describe the framework more generally and create a specific instantiation for the problem of dexterous, in-hand reorientation of a cube by a highly actuated robot hand.
%
\end{abstract}

\section{Introduction}
Imbuing robots with high-performance, real-world manipulation skills is critical for their societal relevance. However, such capabilities remain largely elusive, especially for robots with a large number of actuators and high-frequency joint control. Historically, analytically derived controllers have shown successful demonstrations in enabling robots to reach through their environments \cite{cheng2020rmp, ratliff2018riemannian}, manipulate objects \cite{wimboeck2006passivity}, enable end-effector control \cite{nakanishi2008operational}, and perform insertion \cite{van2018comparative}. Although the behavior of these controllers are well understood along with interpretable parameters, their applicability is limited to our ability to design them which are always accompanied by constraining assumptions.

Recently, there have been an escalating number of advancements in robots learning their manipulation skills. Striking exhibitions include high degree-of-actuation (DoA) robots manipulating objects \cite{andrychowicz2020learning, akkaya2019solving, handa2023dextreme, yin2023rotating} and collaborative arms performing tight-tolerance insertions \cite{tang2023industreal}. Furthermore, these skills were trained entirely in simulation via online RL methods and demonstrate real-world robot manipulation skills that are often not achievable by any other means. Similar findings have been reported in legged robots  \cite{miki2022learning, li2023robust} as well as character animation \cite{peng2022ase, zhang2023learning}.

Reinforcement learning grounded in simulation is a very compelling approach in that it can train value functions and policies across an enormous amount of diverse experience. For instance, decades or centuries of simulated experience can be achieved in only a few real-world days \cite{handa2023dextreme}. However, reinforcement learning does come with challenges like reward engineering, optimization efficacy, and resulting policies are often not safe for real-world deployment. Despite action regularization, RL policies still tend to bang-bang actions which are damaging to actuator drives of robots. To combat these destructive action signals, low-pass filters are often applied to reject high-frequency content in the actions \cite{handa2023dextreme, li2023robust, andrychowicz2020learning, akkaya2019solving, yin2023rotating}, but this still does not handle motor constraints explicitly, can result in sluggish policy behavior, and still admit high-frequency content in actions. 

A promising  avenue forward for sequencing high-performance skills consist of mixing well-understood control frameworks and learning methods. For instance, there exists a growing number of theoretical control frameworks grounded in second-order dynamics like geometric control \cite{bullo2019geometric, susskind2011theoretical}, Dynamic Movement Primitives, \cite{ijspeert2013dynamical}, Riemannian Motion Policies \cite{ratliff2018riemannian, cheng2020rmp, li2019stable}, and most recently, geometric fabrics \cite{van2022geometric, ratliff2023fabrics}. Geometric fabrics generalize classical mechanics and are stable, path consistent, and expressive. We refer the reader to \cite{van2022geometric} for extended discussion on theoretical differences between geometric fabrics and the related control work. All of these methods substantially advance beyond standard joint- or task-space controllers, which typically only enable straightline motions in their targeted spaces \cite{spong2020robot, khatib1987unified, nakanishi2008operational}. Despite this, a significant portion of policy learning is applied over these simplistic controllers \cite{tang2023industreal, dalal2023imitating, mandlekar2021matters, bousmalis2023robocat, brohan2023rt}

Built upon generalized nonlinear geometry \cite{ratliff2021generalized}, geometric fabric policies have shown to outperform RMPs \cite{van2022geometric, xie2023neural}, DMPs \cite{xie2023neural}, and Koopman Operator policies \cite{han2023utility}. RMPs have also been used in RL contexts, leveraging a range of structure imposed by the RMP framework to improve policy performance on reaching tasks \cite{li2021rmp2}. Most recently, an even broader classes of geometric fabrics have been theoretically derived that are easier to construct while retaining all important prior properties \cite{ratliff2023fabrics}. We elect to leverage geometric fabrics in this work because of their empirical performance advancement over prior state-of-the-art methods, their provable stability, and path consistency. Note, RMPs are incredibly broad as defined in \cite{ratliff2018riemannian}, and actually encapsulate geometric fabrics as a special subclass with important properties.

Grounded in these recent advances, this work proposes a general framework that combines RL, second-order control frameworks, and physical dynamics to sequence high-performance manipulation skill. Our contributions include:
\begin{itemize}
    \item A general framework that cascades an RL policy and \textit{behavioral dynamics}: a stacked, second order dynamical system that mixes artificial and real dynamics, shifting the underlying dynamics more favorably.
    \item A quadratic program with closed-form solution that enables acceleration and jerk constraint handling by second-order systems.
    \item A novel geometric fabric of the latest form \cite{ratliff2023fabrics} that handles robot joint position constraints, encourages fingertip contact via geometric paths, and opens a force action space for a highly actuated robot hand.
    \item Instantiates the framework by combining RL, this geometric fabric, and simulation at scale to train dexterous in-hand cube re-orientation skill as described in \cite{handa2023dextreme, andrychowicz2020learning} leading to break-through sim2real performance.
\end{itemize}

\section{Formulation}
\label{sec:Formulation}
We reshape the real second-order dynamics of a robot via an artificial second-order dynamical system. These artificial dynamics are constructed from a recently uncovered family of geometric fabrics \cite{ratliff2023fabrics}. This geometric fabric will generate speed-invariant paths through space, automatically handle certain constraints, capture useful, guiding tendencies, and expose an action space. Policies can issue actions in this space that will mix with the guiding fabric, ultimately generating a combined behavior manifested by the real robot. We call this mixing of artificial and real dynamics, \textit{behavioral dynamics}, which is constructed as follows.

\subsection{Forcing Energized Fabrics}
We leverage recent theory in fabrics and follow a provably stable subclass thereof as described by Theorem IV.1 in \cite{ratliff2023fabrics}. This fabric is the following stable second order dynamical system

\begin{flalign}
\label{eq:fabric}
\qdd_f &= \wt{\h}(\q_f, \qd_f) + \alpha_\Lag (\q_f, \qd_f) \qd_f \\
& -\M_f^{-1}(\q_f, \qd_f) \left(\partial_\psi(\q_f) + \B(\q_f, \qd_f) \qd_f \right) \nonumber \\
& - \beta(\q_f, \qd_f) \qd_f \nonumber
\end{flalign}
where $\q_f, \qd_f, \qdd_f \in \mathbb{R}^n$ are the position, velocity, and acceleration of the fabric with $n$ dimensions. $\M_f \in \mathbb{R}^{n \times n}$ is the positive-definite system metric (mass), which captures system prioritization (dependencies dropped for brevity). $\wt{\h} \in \mathbb{R}^n$ is a fabric, which we make homogeneous of degree 2 in velocity (HD2) to produce geometric paths through space (which can be interpreted as nominal system behavior). $\alpha_\Lag \in \mathbb{R}$ is an energization coefficient which ensures the fabric maintain a certain energy, $\Lag$. $\partial_\psi \in \mathbb{R}^n$ is the gradient of a potential function and $\B \in \mathbb{R}^{n \times n}$ a positive semi-definite damping matrix, both of which additionally perturb system acceleration from the nominal fabric. These can be used to impose constraints on the system, for instance. Finally, $\beta \in \mathbb{R}^+$ is an additional damping scalar that preserves the fabric geometry and serves to stabilize the system by removing energy.

\subsection{Behavioral Dynamics}
The geometric fabric governing the artificial dynamics in (\ref{eq:fabric}) can be compactly rewritten as

\begin{align}
\label{eq:fabrics_dynamics}
    \M_f(\q_f, \qd_f) \qdd_f + \f_f(\q_f, \qd_f) = \mathbf{0}
\end{align}
where $\f_f \in \mathbb{R}^{n}$ is the artificial force. These dynamics are connected to the real dynamics of a robot as

\begin{align}
    \label{eq:real_dynamics}
    \M(\q) \qdd + \f(\q, \qd) = \tau(\q, \qd, \q_f, \qd_f, \qdd_f)
\end{align}
where $\q, \qd, \qdd \in \mathbb{R}^n$ are the real position, velocity, and acceleration. $\M \in \mathbb{R}^{n \times n}$ and $\f \in \mathbb{R}^{n}$ are the real robot mass and force (including contact, Centripetal/Coriolis, friction, and gravity forces). The torque control law, $\tau$, connects the artificial and real dynamics together in some way. A specific instantiation of this torque law is discussed in Section \ref{subsec:torque_controller}.

With the behavioral dynamics above, a policy, $\pi(\cdot)$, can produce a driving force on the fabric by issuing actions $\mathbf{a}$, ($\mathbf{a} \sim \pi(\cdot)$), to some function, $\f_\pi(\cdot)$, as

\begin{align}
\label{eq:fabrics_rl_dynamics}
    \M_f(\q_f, \qd_f) \qdd_f + \f_f(\q_f, \qd_f) + \f_\pi(\mathbf{a}) = \mathbf{0},
\end{align}
where $\f_\pi(\mathbf{a})$ is interpreted as a driving force on the fabric. Consequently, the time evolution of the fabric state, $(\q_f, \qd_f)$, is a function of the fabric itself and the control forces produced by the policy. General control forces over second-order dynamical systems is a ubiquitous control input in many contexts including torque control laws for robots \cite{behal2009lyapunov} and torque control inputs for trajectory optimization \cite{manchester2020variational}. In theory, $\f_\pi(\mathbf{a})$ could destabilize the artificial dynamics, but in practice, sufficiently large $\B$ and $\beta$ maintains system stability. Moreover, energy capping methods (Proposition III.3 in \cite{ratliff2023fabrics}) can guarantee stability even with a general driving force coming from $\f_\pi(\mathbf{a})$. We call $\pi(\cdot)$ a Fabric-Guided Policy (FGP).

\subsection{Torque Controller}
\label{subsec:torque_controller}
Typically, the torque control law in (\ref{eq:real_dynamics}) is a joint-level proportional-derivative (PD) controller with inverse dynamics compensation. This is \textit{not} the only approach to connecting the two dynamical systems, but one that we most often employ. The joint PD route facilitates tracking control which ultimately means that $||\q_f - \q|| \leq \epsilon_1 \in \mathbb{R}^+$ and $||\qd_f - \qd|| \leq \epsilon_2 \in \mathbb{R}^+$. Both $\epsilon_1, \epsilon_2$ can be driven arbitrarily small based how much of the inverse dynamics are compensated and the magnitude of the PD gains (see \cite{behal2009lyapunov} for in-depth analyses). For the particular case of a fully controllable robot arm moving in freespace, one can then see that the geometric fabric effectively replace the real dynamics since $\q \approx \q_f$ and $\qd \approx \qd_f$. More generally, the controllable robot states and the associated fabric state will closely match in free-space and separate during contact. This separation induces contact forces, which can be leveraged to perform mechanical work, e.g., object manipulation. The technique of inducing and controlling contact forces via a separation in target states and desired states is standard practice via impedance and admittance control formulations \cite{ott2010unified, magrini2015control}. This particular separation in fabric and actual state is a useful construction for a variety of reasons. First, it still follows the ubiquitous paradigm of RL policies generating joint actions for an underyling PD controller \cite{miki2022learning, handa2023dextreme, andrychowicz2020learning, peng2022ase, yin2023rotating}. Second, it is also common in Dynamic Motion Primitive \cite{ijspeert2013dynamical}, RMP work \cite{ratliff2018riemannian}, and tight-tolerance insertions with the PLAI scheme in \cite{tang2023industreal} which positively impacted sim2real efforts.

\section{Application to Multi-Fingered Cube Reorientation}
\label{sec:application}
We apply the preceding framework to the problem of in-hand cube orientation by a 16-actuator, 4-finger Allegro Hand v4 with hardware setup and vision-based cube pose estimation exactly as detailed in \cite{handa2023dextreme} (except \cite{handa2023dextreme} used the Allegro Hand v3). This particular problem is quite challenging for many reasons including: 1) hybrid dynamics, 2) complexity of the hand given its geometry, high actuator count, and unmodeled dynamics, and 3) state estimation of the cube. Of course, previous works have surmounted these complexities and have shown strong real-world performance given policies that were trained purely in a physical simulator. We compliment these preceding works by studying the application and effect of our new control framework in this setting and discover important takeaways.

\subsection{Fabrics Design and Policy Action Space}
To realize the architecture covered in Section \ref{sec:Formulation}, we design a forced energized geometric fabric for the Allegro hand following the general form in (\ref{eq:fabric}). The components of this fabric are described as follows. For more information on how these components are combined into (\ref{eq:fabric}), refer to \cite{ratliff2018riemannian, van2022geometric, ratliff2023fabrics}.

\subsubsection{Attraction}
We animate two behavioral elements in the fabric based on manipulation insights. First, encouraging fingertip contact with the cube facilitates its controllability (like strategies in  \cite{van2018comparative}). Second, maintaining inwardly curled fingers around the cube provides caging effects. We imbue the fabric with these behaviors by constructing geometric attractors in two different spaces, $\x$, as $(\M(\x), \xdd(\x, \xd, \x_g)$, where $\x \in \mathbb{R}^{12}$ for the concatenated fingertip space and $\x \in \mathbb{R}^{16}$ for the configuration space. $\M(\x) = m \I$ is a constant isotropic mass, where  $m \in \mathbb{R}^+$. $\xdd = -k_a \|\xd\|^2 \tanh(\alpha_a \|\x - \x_g\|) \frac{\x - \x_g}{\|\x - \x_g \|}$, where $k_a \in \mathbb{R}^+$ is a constant attraction gain, $\alpha_a$ is a constant sharpness parameter, and $\x_g$ is a target state in this space. To engender fingertip contact, we set $\x_g = [\x_c^\tr, \x_c^\tr, \x_c^\tr, \x_c^\tr]^\tr$ ($\x_c \in \mathbb{R}^3$ is the 3D center of the cube) in the fingertip space. To evoke inwardly curling fingers, we set $\x_g$ to some fixed, curled position in the configuration space. These components partially construct the system metric, $\M_f$, and fully construct the geometric fabric, $\wt{\h}$, in (\ref{eq:fabric}).

\subsubsection{Repulsion}
To ensure the fabric state respects the upper and lower joint limits of the robot hand ($\overline{\q}, \underline{\q}$), we introduce repulsion forcing fabric terms in an upper joint limit task space, $\x= \overline{\q} - \q$, and lower joint limit task space, $\x = \q - \underline{\q}$. The metric for the fabric term in these spaces is $\M(\x) = \text{diag} \left( \max(-\text{sgn}(\xd), 0) \frac{k_b}{\x} \right)$, where $k_b \in \mathbb{R}^+$ is a constant gain. Effectively, this is a barrier metric for which a diagonal element $\M_{ii} \to \infty$ as $\x_i \to 0$. The paired acceleration is $\xdd = \mathbf{g} - b \xd$, where $\mathbf{g} \in \mathbb{R}^{n+}$ ($n$ is the dimensionality of the configuration space) is constant and $b \in \mathbb{R}^+$ is a constant damping gain. Effectively, the acceleration is positive with damping. This component partially constructs $\M_f$ in (\ref{eq:fabric}), and fully constructs $\partial_\psi$ and $\B$ in (\ref{eq:fabric}).

\subsubsection{Energization}
The energization coefficient $\alpha_\Lag$ in (\ref{eq:fabric}) is calculated as detailed in Theorem IV.5 in \cite{van2022geometric} from a configuration-space energy, $\Lag = \frac{1}{2}\qd_f^\tr \qd_f$, and the fabric $\wt{\h}$. This ensures that the geometric fabric itself is energy stable.

\subsubsection{Geometrically-Consistent Damping}
The final damping term $\beta$ in (\ref{eq:fabric}) is set to a constant smaller value ($\beta=2.5$) during training to facilitate exploration. During deployment, a variety of $\beta$ levels are tested (see Table \ref{table:experimental_results}, revealing various levels of sim2real performance). Importantly, this damping term influences the speed of resulting movements in a geometrically consistent manner (see \cite{ratliff2021generalized} for more discussion), with higher values incurring slower motion. We found it important to slow the motions down for stronger sim2real transfers as detailed in Section \ref{subsec:real_performance}.

\subsubsection{Acceleration and Jerk Handling}
Robot controllers like the torque law in (\ref{eq:real_dynamics}) typically require that $\q_f$ is sufficiently smooth to protect the actuators, resulting in acceleration and jerk constraints. We can easily accomodate such constraints by formulating the following quadratic program

\begin{align}
    L = \frac{1}{2} (\qdd_f - \qdd)^\tr \M_f (\qdd_f - \qdd) + \frac{\alpha}{2} \qdd_f^\tr \M_f \qdd_f
\end{align}
where $\alpha \in \mathbb{R}^+$ effectively regularizes $||\qdd|| \to 0$ while considering $\M_f$. The closed-form solution is

\begin{align}
    (\M_f + \alpha \mathbf{I}) \qdd_f + \f_f = \mathbf{0}
\end{align}
where $\f_f = -\M_f \qdd$. Solving for $\qdd_f$ produces $\qdd_f = -(\M_f + \alpha \mathbf{I})^{-1} \f_f$ and we can see that as $\alpha \to \infty$, $||\qdd_f|| \to 0$. This means that accelerations can be made arbitrarily small on-demand and we can drive them under the acceleration limits. That is, we can solve for a single $\alpha$ such that $|\qdd_{f, i}| \leq \overline{\qdd}_i \: \forall \: i$, where $\overline{\qdd}_i$ is the $i^{th}$ joint acceleration limit.

We can easily extend the above to accommodate joint jerk limits as well via the following time-discretized jerk model (superscripts indicate time index)

\begin{align}
    \dddot{\mathbf{q}}_f^{t} = \frac{\qdd_f^{t+1} - \qdd_f^t}{\Delta t}
\end{align}
The largest possible jerk will occur when the next acceleration is the maximum acceleration and the previous acceleration is the minimum acceleration (assuming $\underline{\dddot{\mathbf{q}}}_f^t = -\overline{\dddot{\mathbf{q}}}_f^t$), or

\begin{align}
    \overline{\dddot{\mathbf{q}}}_f^{t} = \frac{2\overline{\qdd}}{\Delta t}
\end{align}
Therefore, if jerk must not exceed some limit, $\overline{\dddot{\mathbf{q}}}$, then
\begin{align}
    \frac{2\overline{\qdd}}{\Delta t} \leq = \overline{\dddot{\mathbf{q}}}.
\end{align}
Therefore, we can calculate a single acceleration limit 

\begin{align}
\overline{\qdd} = \text{min} \left( \overline{\qdd}, \frac{\Delta t \overline{\dddot{\mathbf{q}}}}{2\overline{\qdd}} \right)
\end{align}
that respects both the original acceleration limit and jerk limit. With this new $\overline{\qdd}$, we can run the previously detailed scheme for ensuring that both acceleration and jerk constraints are upheld.

\subsubsection{Action Space}
We elect for the RL policy actions to be converted to forces in the concatenated fingertip space resulting in $\mathbf{a} \in \mathbb{R}^{12}$. This force is created by first clamping the actions between $[-1, 1]$ and then scaling them by some positive factor, $\gamma \in \mathbb{R}^+$. This conversion becomes a force in the fingertip space, which is pulled-back to the root of the fabric via this map's Jacobian, i.e., $\f_\pi(\mathbf{a}) = \gamma \J^\tr(\q_f) \text{clamp} (\mathbf{a}, -1, 1)$ in (\ref{eq:fabrics_rl_dynamics}). Interestingly, this RL action space is radically different and results in high policy performance. 

\subsubsection{Numerical Integration}
After evaluating the policy force actions $\f_\pi$ and fabric in (\ref{eq:fabrics_rl_dynamics}), the resulting fabric acceleration at time $t$, $\qdd_f^t$, is forward integrated with an approximate RK2 integration scheme as in \cite{gruver2022deconstructing}. This scheme calculates the next fabric joint position and velocity, $\q_f^{t+1}$ and $\qd_f^{t+1}$, from the current fabric joint position and velocity, $\q_f^t$ and $\qd_f^t$, acceleration $\qdd_f^t$, and timestep $\Delta t$ as
\begin{flalign}
\q_f^{t+1} &= \q_f^t + \Delta t \qd_f^t + \frac{1}{2} \Delta t ^2 \qdd_f^t \\
\qd_f^{t+1} &= \qd_f^t + \Delta t \qdd_f^t
\end{flalign}
Policies issue actions at 30 Hz, while the fabric is forward integrated at 60 Hz, resulting in an integration timestep of $\Delta t = \frac{1}{60}$. The fabric position and velocities are passed as inputs to the torque law in \ref{eq:real_dynamics}, which is a PD controller with gains $k_p=2$ and $k_d=0.1$.

\subsection{Reinforcement Learning Setup}
\label{subsec:rl_setup}
We keep the exact same reward terms and their weights as in \cite{handa2023dextreme}, but completely remove all penalties (action, action delta, and joint velocity penalties) since the fabric layer ensures smooth, safe motion that is within the hardware constraints of the robot. Thus, the fabric layer simplifies reward engineering as well. The neural architectures and sizes for both the value function and policy are the same as in \cite{handa2023dextreme}, with the single exception that both $\q_f$ and $\qd_f$ are additionally given as inputs to both the policy and value function. We also use PPO for RL training with the same hyperparameters and automatic domain randomization (ADR) setup in \cite{handa2023dextreme}, and trained with eight NVIDIA A40 GPUs for about 170 hours of wall-clock time.

\subsection{Cube Disturbance Wrench}
\label{subsec:disturbance_wrench}
We found that policies trained with the force disturbance in \cite{handa2023dextreme} often resulted in macro-level behaviors that did not transfer well to the real world. For instance, fingers would full extend at times allowing the cube to roll out of the hand in the real world. To close this sim2real gap, we applied a full wrench disturbance to the cube during training which forced the policies to more frequently establish prehensile-lock on the cube to maintain controllability over the more chaotic cube dynamics. This change ultimately generated significantly better sim2real performance.

For every action step, a new disturbance wrench expressed and applied in cube-centric coordinates, $C$, is formed (per environment) with 10 \% chance as

\begin{align}
    ^{C}\mathbf{w}_d =
    \begin{bmatrix}
    ^{C}\mathbf{f}_d \\
    ^{C}\mathbf{\tau}_d
    \end{bmatrix} =
    \begin{bmatrix}
    c_1 m ^{C} \hat{\f}_d \\
    ^{C} \mathbf{r} \times ^{C} \mathbf{f}_d
    \end{bmatrix} =
    \begin{bmatrix}
    c_1 m ^{C} \hat{\f}_d \\
    c_2 ^{C} \hat{\mathbf{r}} \times c_1 m ^{C} \hat{\f}_d    
    \end{bmatrix}
\end{align}
where $c_1 \in \mathbb{R}^+$ is an acceleration constant, $m \in \mathbb{R}^+$ is the randomized mass of the cube, and $^{C} 
 \hat{\f}_d \in \mathbb{R}^3$ is randomly sampled direction of unit magnitude. Finally, $c_2 = \frac{\sqrt{3}}{2} 0.065$ is the radius of a sphere centered with the cube (cube side length is 0.065 m), and $^{C}  \hat{\mathbf{r}} \in \mathbb{R}^3$ is a random direction emanating from the origin of the cube. If a new disturbance wrench is not sampled by chance, then the previously sampled wrench is applied. Maintaining a consistent disturbance wrench for several time steps allows for greater influence over the cube's motion. 

\subsection{Training Performance}
We train five random seeds of both FGP and DeXtreme policies on the training setup as described in \cite{handa2023dextreme} with a few modifications. For the FGPs, we augment the inputs to the policy and value function as described in Sections \ref{subsec:rl_setup} and also apply the new disturbance wrench as in \ref{subsec:disturbance_wrench}. For the Dextreme policies, we apply the new disturbance wrench (called DeXtreme (new)). As shown in Table \ref{table:training_results}, DeXtreme policies train faster than FGP policies, particularly in the amount of time required to achieve -0.5 nats per dimension (npd) as defined in \cite{handa2023dextreme}. FGPs also produce lower entropy levels than DeXtreme policies in the allotted training time. Interestingly, neither the FGPs or DeXtreme policies are converged after 170 hours of training and higher levels of entropy could be obtained if trained for longer. Overall, it is harder to learn smoother policies with RL as it effects exploration \cite{seyde2021bang}. However, high-performing FGP policies can still be trained in about 1 week.

\begin{table}
\vspace{2mm}
\centering
\resizebox{1.\linewidth}{!}{
\begin{tabular}{c|c|c|c|c}
\hline 
\textbf{Policy} & \textbf{Seed} & \textbf{Time to -1.0 npd (hours)} & \textbf{Time to -0.5 npd (hours)} & \textbf{Final npd}\\ \hline
\multirow{5}{*}{FGP (ours)}
& 1 & 26.7 & 113.6 & -2.1e-1 \\
& 2 & 17.3 & 139.1 & -3.2e-1 \\
& 3 & 35.6 & N/A & -9.3e-1 \\
& 4 & 20.84 & 121.567 & -2.2e-1 \\
& 5 & 26.5 & 131.457 & -2.6e-1 \\
\hline
\multirow{1}{*}{DeXtreme (new)}
& 1 & 16.6 & 38.4 & -8.0e-2 \\
& 2 & 15.7 & 32.5 & 1.8e-3 \\
& 3 & 31.3 & 52.6 & 3.0e-2 \\
& 4 & 15.4 & 51.7 & -8.2e-3 \\
& 5 & 15.9 & 33.2 & 4.7e-3 \\
\hline
\end{tabular}
}
\caption{Entropy levels obtained by both FGP and DeXtreme policies across random seeds. }
\label{table:training_results}
\vspace{-3mm}
\end{table}

\subsection{Real-World Performance}
\label{subsec:real_performance}
We analyze policy performance across three performance metrics: consecutive success (CS), rotations per minute (RPM) (see Table \ref{table:experimental_results}), and action noise rejection. CS is an established metric for this task and counts the consecutively successful rotations with a rotational goal tolerance of 0.4 rad in the real world \cite{andrychowicz2020learning, handa2023dextreme}. RPM is the CS for a run divided by the duration of the run measured in minutes. Overall, CS captures reliability of the manipulation skill and RPM captures its solution speed. Finally, action noise rejection depicts the level of attenuation present in the action signals in excess of 5 Hz.

The top two highest performing policies along the CS index are the FGP with $\beta=40$ and the DeXtreme (new) policy with an integrator in the torque law. Interestingly, the FGP policy improves with increasing damping levels up to $\beta=40$, after which, performance degrades. We believe this performance trend is due to unmodeled dynamics effects in the simulator, latency effects, and the fact that $\beta$ damps the fabric in a geometrically consistent manner (finger paths are relatively unchanged with changes in $\beta$). The DeXtreme (new) policy had two runs with over 600 CS which significantly increased the mean CS. However, the resulting median of 70 was less than that of the FGP. Since elements of the physical setup have changed since DeXtreme \cite{handa2023dextreme} (new hand, motors and EMA of 0.05 versus 0.1), we re-evaluated the original DeXtreme policy. Despite these changes, we found that the original DeXtreme policy performed consistently (e.g., CS mean of 29.0 vs 27.8). Ultimately, both DeXtreme (new) and FGP policies are very high performing with several runs well over 100 CS, significantly surpassing all prior results  \cite{andrychowicz2020learning,  akkaya2019solving, handa2023dextreme}.

Along the RPM index, the FGP policy significantly outperforms all DeXtreme policies. Interestingly, the RPMs across $\beta$ levels do not fluctuate much and higher $\beta$ leads to greater precision along this metric. From observation, we see that finger movement is indeed faster with smaller $\beta$ levels (as expected), but RPMs do not improve because more manipulation errors are present. Qualitatively, the policies look more meticulous with increased $\beta$, and quantitatively, enable greater precision in solve rates and higher CS. The DeXtreme policies had much lower RPMs, likely due to the EMA filter possessing a much smaller cut-off frequency than what is set during training (EMA factor of 0.15 versus 0.05). Small EMA factors stifle exploration and learning progress during RL training, so larger values are used during training. 

\begin{table*}[htp]
\vspace{2mm}
\centering
\resizebox{.8\linewidth}{!}{
\begin{tabular}{c|c|c|c|c|c|c}
\hline 
\textbf{Policy} & \textbf{$\beta$} & \textbf{CS Mean} & \textbf{CS Median} & \textbf{RPM Mean} & \textbf{RPM Median} & \textbf{CS capped at 50}\\ \hline
\multirow{6}{*}{FGP}
& 2.5 & 24.9 $\pm$ 11.4 & 20.0 & 8.9 $\pm$ 1.0 & \textbf{9.3} & 22.2 $\pm$ 7.8\\
& 10 & 35.4 $\pm$ 17.1 & 25.0 & 8.3 $\pm$ 1.2 & 9.0 & 27.7 $\pm$ 9.5\\
& 20 & 32.6 $\pm$ 14.0 & 29.5 & \textbf{9.9 $\pm$ 1.3} & 9.1 & 28.3 $\pm$ 8.7\\
& 30 & 57.6 $\pm$ 22.1 & 67.0 & 9.0 $\pm$ 0.5 & 9.1 & 36.4 $\pm$ 9.7\\
& 40 & \textbf{94.1 $\pm$ 29.4} & \textbf{85.5} & \textbf{9.4 $\pm$ 0.3} & \textbf{9.4} & \textbf{47.4 $\pm$ 3.1}\\
& 50 & 79.6 $\pm$ 31.1 & 57.5 & 8.7 $\pm$ 0.5 & 8.9 & \textbf{43.0 $\pm$ 6.6} \\ \hline
\multirow{1}{*}{DeXtreme* (new)}
& N/A & \textbf{244.6 $\pm$ 140.1} & \textbf{70.0} & 7.6 $\pm$ 0.4 & 7.6 & 42.3 $\pm$ 7.7\\ \hline
\multirow{1}{*}{DeXtreme* (previous, rerun) \cite{handa2023dextreme}}
& N/A & 29.0 $\pm$ 18.7 & 15.5 & 6.5 $\pm$ 1.6 & 5.7 & 21.1 $\pm$ 10.2 \\ \hline
\multirow{1}{*}{DeXtreme* (previous) \cite{handa2023dextreme}}
& N/A & 27.8 $\pm$ 19.0 & 14.0 & N/A & N/A & 23.1 $\pm$ 9.4\\ \hline
\multirow{1}{*}{OpenAI \cite{andrychowicz2020learning}}
& N/A & 15.2 $\pm$ 14.3 & 11.5 & N/A & N/A & 15.2 $\pm$ 14.3\\ \hline
\vspace{-5mm}
\end{tabular}
}
\caption{Consecutive success (CS) and rotations per minute (RPM) metrics for vision-based policies across 10 runs ($\pm$ indicates 95 \% confidence interval following a t-distribution.) (*) Indicates integrator present in torque law during deployment. }
\label{table:experimental_results}
\end{table*}

\begin{figure}
\vspace{2mm}
\centering
\includegraphics[width=.45\textwidth]{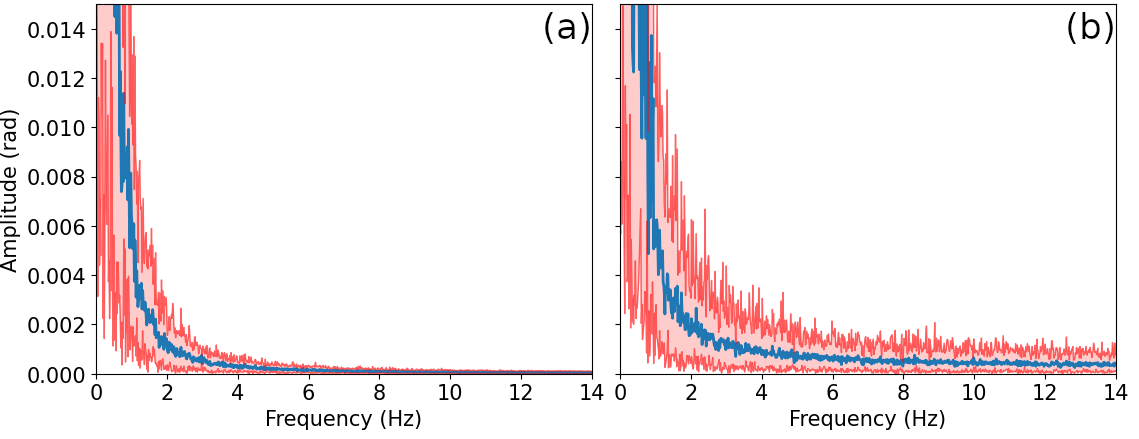}
\caption{Spectral content of target joint angles generated by the (a) FGP and (b) DeXtreme policies indicating greater noise attenuation in FGP control. Blue curve is mean amplitude, red curves are the minimum and maximum amplitudes.}
\label{fig:policy_fft}
\vspace{-3mm}
\end{figure}

Finally, policies should not generate unnecessary noise in their actions as this can accelerate wear-and-tear on real robot hardware. Some suggest frequencies under 5 Hz, covering a large spectrum of robotics applications \cite{lutter2021value}). We inspect the spectral content of the target joint angles produced by the FGP at $\beta=40$ and DeXtreme (new) policies by passing a one minute recording of these signals through the Fast-Fourier Transform (FFT) per joint (see Fig. \ref{fig:policy_fft}). As shown, most of the action signals emit frequencies with high amplitudes below 2 Hz for both policies. However, the DeXtreme policy produces significantly higher magnitude action noise in excess of 5 Hz, indicating stronger noise rejection capabilities of the FGP. This is particularly interesting considering that the raw RL actions $\mathbf{a}$ in (\ref{eq:fabrics_rl_dynamics}) for the fabric are \textit{bang-bang}. Smoothness should be adopted as a critical performance index for RL policies as they become more widely adopted with significant implications on the running costs of hardware maintenance and loss of productivity for repairs.

\section{Discussion}
\label{sec:discussion}
Training an RL policy over a fabric-based behavioral dynamics layer engendered new state-of-the-art performance for in-hand cube re-orientation. The FGP at $\beta=40$ more than tripled performance along the CS metric when compared to the previous state-of-the-art DeXtreme model. Against the latest DeXtreme model, the FGP had significantly lower mean CS, but higher median CS. The FGP also had the fastest solve times with the highest and most consistent RPM. Finally, the FGP produced very little action noise, if any, with spectral amplitudes nearly zero above 5 Hz, whereas DeXtreme policies still admitted frequencies above 5 Hz despite its heavy usage of low-pass filtering. Note, we do not focus on generalizing across object geometries as in \cite{chen2022system, chen2022visual}, but instead, focus on generalizing across other aspects of system dynamics and searching for methods that push skill performance towards industrial-grade levels.

In general, the sim2real transfers among FGP and DeXtreme policies are variable despite high ADR entropy and CS among seeds in simulation. We believe domain randomization is a necessary but insufficient condition for maximizing real-world performance. In reality, different training runs evolve different macro-level behaviors. Since the reward functions under-specify the desired behavior and optimization is also subject to many local minima, policies can converge onto a continuum of different strategies for solving the problem in simulation. However, some strategies transfer more strongly, a phenomenon also expressed in \cite{handa2023dextreme, escontrela2022adversarial}. For example, policies that tended to cage the cube more often transferred more strongly than policies that too frequently fully extended the fingers, allowing the opportunity for the cube to roll out of the hand. Training DeXtreme policies for much longer than in \cite{handa2023dextreme} resulted in much higher ADR entropy and very high CS in the real world, providing evermore credence towards our infrastructure, approach, and large-scale training runs.  Interestingly, none of the FPG and DeXtreme policies ever converged during RL training. Further increasing the duration of the training runs could amplify real-world performance even more as the policies continue to improve their generalization across system dynamics.

Smoother policies impede RL progression and counters RL's preference towards bang-bang control \cite{seyde2021bang}. This was observed in DeXtreme, which is the reason for higher EMA factors during training. Similarly, FGP policies train more slowly than DeXtreme, which is very likely due to their even smoother action profiles (see Section \ref{subsec:real_performance} for details). In general, advancements need to be made to improve RL for smooth policies which could include automatically optimizing over various hyperparameters \cite{petrenko2023dexpbt}, employing more effective RL algorithms, investigating alternative exploration noise \cite{eberhard2022pink}, or leveraging priors in some form \cite{escontrela2022adversarial}.

Designing and tuning a geometric fabric does require deep understanding of the control method, experience, and vectorized tooling to enable training at scale. There is a rich history of practitioners succeeding at designing these second order systems \cite{ratliff2018riemannian, van2022geometric, cheng2020rmp, li2019stable, li2021rmp2} and the recent simplifications to geometric fabrics \cite{ratliff2023fabrics} further ease the process. Moreover, the fabric terms themselves are not overly exotic and are often very compact and interpretable equations as shown in Section \ref{sec:application}. With the right software tooling, an experienced practitioner can design and tune a fabric within a few hours, which is significantly faster than iterating over more complicated reward functions and extended RL training runs. Besides, optimizers always have the burden of decoding more complex reward functions directly, resulting in local minima and deficiencies in optimization fidelity. Instead, fabrics can directly encode behavior in closed-form, simplifying reward engineering and providing stronger guarantees on some behavioral elements. For instance, the geometric fabric herein upheld joint position, acceleration, and jerk constraints. The FGP directly benefits from the inherent constraint handling of the fabric and does not require additional low-pass filtering, enabling fast and safe motion. Additionally, the fabric generated fingertip paths towards the cube, guiding the RL policy to more strongly leverage the fingertips during manipulation. The geometric nature of this tendency also enabled deploying the fabric at different levels of $\beta$, allowing its optimization for maximizing real-world performance. Note, the design space for geometric fabrics is very large and different designs can influence FGP performance.

\section{Conclusion and Future Work}
\label{sec:conclusion}
We propose a general paradigm for cascading an RL policy, an artificial dynamical system, and real system dynamics together. We instantiate this paradigm with a powerful second-order control method, geometric fabrics, reinforcement learning, and vectorized simulation at scale to achieve state-of-the-art policy performance for the established cube re-orientation task by a dexterous, high DoA robot hand. Future work includes applying the approach more broadly across different robot platforms, different tasks, and different fabric designs. We will also consider other optimization and planning methods in search of maximizing policy performance.

{
\small
\bibliographystyle{IEEEtran}
\bibliography{refs}

\begin{thebibliography}{10}
\providecommand{\url}[1]{#1}
\csname url@samestyle\endcsname
\providecommand{\newblock}{\relax}
\providecommand{\bibinfo}[2]{#2}
\providecommand{\BIBentrySTDinterwordspacing}{\spaceskip=0pt\relax}
\providecommand{\BIBentryALTinterwordstretchfactor}{4}
\providecommand{\BIBentryALTinterwordspacing}{\spaceskip=\fontdimen2\font plus
\BIBentryALTinterwordstretchfactor\fontdimen3\font minus
  \fontdimen4\font\relax}
\providecommand{\BIBforeignlanguage}[2]{{%
\expandafter\ifx\csname l@#1\endcsname\relax
\typeout{** WARNING: IEEEtran.bst: No hyphenation pattern has been}%
\typeout{** loaded for the language `#1'. Using the pattern for}%
\typeout{** the default language instead.}%
\else
\language=\csname l@#1\endcsname
\fi
#2}}
\providecommand{\BIBdecl}{\relax}
\BIBdecl

\bibitem{cheng2020rmp}
C.-A. Cheng, M.~Mukadam, J.~Issac, S.~Birchfield, D.~Fox, B.~Boots, and
  N.~Ratliff, ``Rmp flow: A computational graph for automatic motion policy
  generation,'' in \emph{Algorithmic Foundations of Robotics XIII: Proceedings
  of the 13th Workshop on the Algorithmic Foundations of Robotics 13}.\hskip
  1em plus 0.5em minus 0.4em\relax Springer, 2020, pp. 441--457.

\bibitem{ratliff2018riemannian}
N.~D. Ratliff, J.~Issac, D.~Kappler, S.~Birchfield, and D.~Fox, ``Riemannian
  motion policies,'' \emph{arXiv preprint arXiv:1801.02854}, 2018.

\bibitem{wimboeck2006passivity}
T.~Wimboeck, C.~Ott, and G.~Hirzinger, ``Passivity-based object-level impedance
  control for a multifingered hand,'' in \emph{2006 IEEE/RSJ International
  Conference on Intelligent Robots and Systems}.\hskip 1em plus 0.5em minus
  0.4em\relax IEEE, 2006, pp. 4621--4627.

\bibitem{nakanishi2008operational}
J.~Nakanishi, R.~Cory, M.~Mistry, J.~Peters, and S.~Schaal, ``Operational space
  control: A theoretical and empirical comparison,'' \emph{The International
  Journal of Robotics Research}, vol.~27, no.~6, pp. 737--757, 2008.

\bibitem{van2018comparative}
K.~Van~Wyk, M.~Culleton, J.~Falco, and K.~Kelly, ``Comparative peg-in-hole
  testing of a force-based manipulation controlled robotic hand,'' \emph{IEEE
  Transactions on Robotics}, vol.~34, no.~2, pp. 542--549, 2018.

\bibitem{andrychowicz2020learning}
O.~M. Andrychowicz, B.~Baker, M.~Chociej, R.~Jozefowicz, B.~McGrew,
  J.~Pachocki, A.~Petron, M.~Plappert, G.~Powell, A.~Ray \emph{et~al.},
  ``Learning dexterous in-hand manipulation,'' \emph{The International Journal
  of Robotics Research}, vol.~39, no.~1, pp. 3--20, 2020.

\bibitem{akkaya2019solving}
I.~Akkaya, M.~Andrychowicz, M.~Chociej, M.~Litwin, B.~McGrew, A.~Petron,
  A.~Paino, M.~Plappert, G.~Powell, R.~Ribas \emph{et~al.}, ``Solving rubik's
  cube with a robot hand,'' \emph{arXiv preprint arXiv:1910.07113}, 2019.

\bibitem{handa2023dextreme}
A.~Handa, A.~Allshire, V.~Makoviychuk, A.~Petrenko, R.~Singh, J.~Liu,
  D.~Makoviichuk, K.~Van~Wyk, A.~Zhurkevich, B.~Sundaralingam \emph{et~al.},
  ``De{X}treme: Transfer of agile in-hand manipulation from simulation to
  reality,'' in \emph{2023 IEEE International Conference on Robotics and
  Automation (ICRA)}.\hskip 1em plus 0.5em minus 0.4em\relax IEEE, 2023, pp.
  5977--5984.

\bibitem{yin2023rotating}
Z.-H. Yin, B.~Huang, Y.~Qin, Q.~Chen, and X.~Wang, ``Rotating without seeing:
  Towards in-hand dexterity through touch,'' \emph{arXiv preprint
  arXiv:2303.10880}, 2023.

\bibitem{tang2023industreal}
B.~Tang, M.~A. Lin, I.~Akinola, A.~Handa, G.~S. Sukhatme, F.~Ramos, D.~Fox, and
  Y.~Narang, ``Industreal: Transferring contact-rich assembly tasks from
  simulation to reality,'' \emph{Robotics: Science and Systems}, 2023.

\bibitem{miki2022learning}
T.~Miki, J.~Lee, J.~Hwangbo, L.~Wellhausen, V.~Koltun, and M.~Hutter,
  ``Learning robust perceptive locomotion for quadrupedal robots in the wild,''
  \emph{Science Robotics}, vol.~7, no.~62, p. eabk2822, 2022.

\bibitem{li2023robust}
Z.~Li, X.~B. Peng, P.~Abbeel, S.~Levine, G.~Berseth, and K.~Sreenath, ``Robust
  and versatile bipedal jumping control through multi-task reinforcement
  learning,'' \emph{arXiv preprint arXiv:2302.09450}, 2023.

\bibitem{peng2022ase}
X.~B. Peng, Y.~Guo, L.~Halper, S.~Levine, and S.~Fidler, ``Ase: Large-scale
  reusable adversarial skill embeddings for physically simulated characters,''
  \emph{ACM Transactions On Graphics (TOG)}, vol.~41, no.~4, pp. 1--17, 2022.

\bibitem{zhang2023learning}
H.~Zhang, Y.~Yuan, V.~Makoviychuk, Y.~Guo, S.~Fidler, X.~B. Peng, and
  K.~Fatahalian, ``Learning physically simulated tennis skills from broadcast
  videos,'' \emph{ACM Transactions on Graphics (TOG)}, vol.~42, no.~4, pp.
  1--14, 2023.

\bibitem{bullo2019geometric}
F.~Bullo and A.~D. Lewis, \emph{Geometric control of mechanical systems:
  modeling, analysis, and design for simple mechanical control systems}.\hskip
  1em plus 0.5em minus 0.4em\relax Springer, 2019, vol.~49.

\bibitem{susskind2011theoretical}
L.~Susskind, ``The theoretical minimum: Classical mechanics,'' 2011.

\bibitem{ijspeert2013dynamical}
A.~J. Ijspeert, J.~Nakanishi, H.~Hoffmann, P.~Pastor, and S.~Schaal,
  ``Dynamical movement primitives: learning attractor models for motor
  behaviors,'' \emph{Neural computation}, vol.~25, no.~2, pp. 328--373, 2013.

\bibitem{li2019stable}
A.~Li, C.-A. Cheng, B.~Boots, and M.~Egerstedt, ``Stable, concurrent controller
  composition for multi-objective robotic tasks,'' in \emph{2019 IEEE 58th
  Conference on Decision and Control (CDC)}.\hskip 1em plus 0.5em minus
  0.4em\relax IEEE, 2019, pp. 1144--1151.

\bibitem{van2022geometric}
K.~Van~Wyk, M.~Xie, A.~Li, M.~A. Rana, B.~Babich, B.~Peele, Q.~Wan, I.~Akinola,
  B.~Sundaralingam, D.~Fox \emph{et~al.}, ``Geometric fabrics: Generalizing
  classical mechanics to capture the physics of behavior,'' \emph{IEEE Robotics
  and Automation Letters}, vol.~7, no.~2, pp. 3202--3209, 2022.

\bibitem{ratliff2023fabrics}
N.~D. Ratliff and K.~Van~Wyk, ``Fabrics: a foundationally stable medium for
  encoding prior experience,'' \emph{arXiv preprint}, 2023.

\bibitem{spong2020robot}
M.~W. Spong, S.~Hutchinson, and M.~Vidyasagar, \emph{Robot modeling and
  control}.\hskip 1em plus 0.5em minus 0.4em\relax John Wiley \& Sons, 2020.

\bibitem{khatib1987unified}
O.~Khatib, ``A unified approach for motion and force control of robot
  manipulators: The operational space formulation,'' \emph{IEEE Journal on
  Robotics and Automation}, vol.~3, no.~1, pp. 43--53, 1987.

\bibitem{dalal2023imitating}
M.~Dalal, A.~Mandlekar, C.~Garrett, A.~Handa, R.~Salakhutdinov, and D.~Fox,
  ``Imitating task and motion planning with visuomotor transformers,''
  \emph{arXiv preprint arXiv:2305.16309}, 2023.

\bibitem{mandlekar2021matters}
A.~Mandlekar, D.~Xu, J.~Wong, S.~Nasiriany, C.~Wang, R.~Kulkarni, L.~Fei-Fei,
  S.~Savarese, Y.~Zhu, and R.~Mart{\'\i}n-Mart{\'\i}n, ``What matters in
  learning from offline human demonstrations for robot manipulation,''
  \emph{arXiv preprint arXiv:2108.03298}, 2021.

\bibitem{bousmalis2023robocat}
K.~Bousmalis, G.~Vezzani, D.~Rao, C.~Devin, A.~X. Lee, M.~Bauza, T.~Davchev,
  Y.~Zhou, A.~Gupta, A.~Raju \emph{et~al.}, ``Robocat: A self-improving
  foundation agent for robotic manipulation,'' \emph{arXiv preprint
  arXiv:2306.11706}, 2023.

\bibitem{brohan2023rt}
A.~Brohan, N.~Brown, J.~Carbajal, Y.~Chebotar, X.~Chen, K.~Choromanski,
  T.~Ding, D.~Driess, A.~Dubey, C.~Finn \emph{et~al.}, ``Rt-2:
  Vision-language-action models transfer web knowledge to robotic control,''
  \emph{arXiv preprint arXiv:2307.15818}, 2023.

\bibitem{ratliff2021generalized}
N.~D. Ratliff, K.~Van~Wyk, M.~Xie, A.~Li, and M.~A. Rana, ``Generalized
  nonlinear and finsler geometry for robotics,'' in \emph{2021 IEEE
  International Conference on Robotics and Automation (ICRA)}.\hskip 1em plus
  0.5em minus 0.4em\relax IEEE, 2021, pp. 10\,206--10\,212.

\bibitem{xie2023neural}
M.~Xie, A.~Handa, S.~Tyree, D.~Fox, H.~Ravichandar, N.~D. Ratliff, and
  K.~Van~Wyk, ``Neural geometric fabrics: Efficiently learning high-dimensional
  policies from demonstration,'' in \emph{Conference on Robot Learning}.\hskip
  1em plus 0.5em minus 0.4em\relax PMLR, 2023, pp. 1355--1367.

\bibitem{han2023utility}
Y.~Han, M.~Xie, Y.~Zhao, and H.~Ravichandar, ``On the utility of koopman
  operator theory in learning dexterous manipulation skills,'' \emph{arXiv
  preprint arXiv:2303.13446}, 2023.

\bibitem{li2021rmp2}
A.~Li, C.-A. Cheng, M.~A. Rana, M.~Xie, K.~Van~Wyk, N.~Ratliff, and B.~Boots,
  ``Rmp2: A structured composable policy class for robot learning,''
  \emph{Robotics: Science and Systems}, 2021.

\bibitem{behal2009lyapunov}
A.~Behal, W.~Dixon, D.~M. Dawson, and B.~Xian, \emph{Lyapunov-based control of
  robotic systems}.\hskip 1em plus 0.5em minus 0.4em\relax CRC Press, 2009,
  vol.~36.

\bibitem{manchester2020variational}
Z.~Manchester and S.~Kuindersma, ``Variational contact-implicit trajectory
  optimization,'' in \emph{Robotics Research: The 18th International Symposium
  ISRR}.\hskip 1em plus 0.5em minus 0.4em\relax Springer, 2020, pp. 985--1000.

\bibitem{ott2010unified}
C.~Ott, R.~Mukherjee, and Y.~Nakamura, ``Unified impedance and admittance
  control,'' in \emph{2010 IEEE international conference on robotics and
  automation}.\hskip 1em plus 0.5em minus 0.4em\relax IEEE, 2010, pp. 554--561.

\bibitem{magrini2015control}
E.~Magrini, F.~Flacco, and A.~De~Luca, ``Control of generalized contact motion
  and force in physical human-robot interaction,'' in \emph{2015 IEEE
  international conference on robotics and automation (ICRA)}.\hskip 1em plus
  0.5em minus 0.4em\relax IEEE, 2015, pp. 2298--2304.

\bibitem{gruver2022deconstructing}
N.~Gruver, M.~Finzi, S.~Stanton, and A.~G. Wilson, ``Deconstructing the
  inductive biases of hamiltonian neural networks,'' \emph{arXiv preprint
  arXiv:2202.04836}, 2022.

\bibitem{seyde2021bang}
T.~Seyde, I.~Gilitschenski, W.~Schwarting, B.~Stellato, M.~Riedmiller,
  M.~Wulfmeier, and D.~Rus, ``Is bang-bang control all you need? solving
  continuous control with bernoulli policies,'' \emph{Advances in Neural
  Information Processing Systems}, vol.~34, pp. 27\,209--27\,221, 2021.

\bibitem{lutter2021value}
M.~Lutter, S.~Mannor, J.~Peters, D.~Fox, and A.~Garg, ``Value iteration in
  continuous actions, states and time,'' \emph{arXiv preprint
  arXiv:2105.04682}, 2021.

\bibitem{chen2022system}
T.~Chen, J.~Xu, and P.~Agrawal, ``A system for general in-hand object
  re-orientation,'' in \emph{Conference on Robot Learning}.\hskip 1em plus
  0.5em minus 0.4em\relax PMLR, 2022, pp. 297--307.

\bibitem{chen2022visual}
T.~Chen, M.~Tippur, S.~Wu, V.~Kumar, E.~Adelson, and P.~Agrawal, ``Visual
  dexterity: In-hand dexterous manipulation from depth,'' 2022.

\bibitem{escontrela2022adversarial}
A.~Escontrela, X.~B. Peng, W.~Yu, T.~Zhang, A.~Iscen, K.~Goldberg, and
  P.~Abbeel, ``Adversarial motion priors make good substitutes for complex
  reward functions,'' in \emph{2022 IEEE/RSJ International Conference on
  Intelligent Robots and Systems (IROS)}.\hskip 1em plus 0.5em minus
  0.4em\relax IEEE, 2022, pp. 25--32.

\bibitem{petrenko2023dexpbt}
A.~Petrenko, A.~Allshire, G.~State, A.~Handa, and V.~Makoviychuk, ``Dex{PBT}:
  Scaling up dexterous manipulation for hand-arm systems with population based
  training,'' \emph{Robotics: Science and Systems}, 2023.

\bibitem{eberhard2022pink}
O.~Eberhard, J.~Hollenstein, C.~Pinneri, and G.~Martius, ``Pink noise is all
  you need: Colored noise exploration in deep reinforcement learning,'' in
  \emph{The Eleventh International Conference on Learning Representations},
  2023.

\end{thebibliography}
}




\end{document}